\documentclass[sigconf]{acmart}

\usepackage{booktabs} 
\usepackage{multirow}
\usepackage{graphicx}
\usepackage{subcaption}
\usepackage{booktabs}
\usepackage{makecell}
\usepackage{lipsum}
\usepackage{algorithm}
\usepackage{algpseudocode}

\title{Semantic Smoothing via Novel View Synthesis \\ for Robust SAR Image Classification}

\author{Daniel Brignac}
\affiliation{%
  \institution{The University of Arizona}
  \city{Tucson, Arizona}
  \country{United States}
}
\email{dbrignac@arizona.edu }

\author{Fengwei Tian}
\affiliation{%
  \institution{The University of Arizona}
  \city{Tucson, Arizona}
  \country{United States}
}
\email{fengtian@arizona.edu}

\author{Banafsheh Latibari}
\affiliation{%
  \institution{The University of Arizona}
  \city{Tucson, Arizona}
  \country{United States}
}
\email{banafsheh@arizona.edu}

\author{Abhijit Mahalanobis}
\affiliation{%
  \institution{The University of Arizona}
  \city{Tucson, Arizona}
  \country{United States}
}
\email{amahalan@arizona.edu}

\author{Ravi Tandon}
\affiliation{%
  \institution{The University of Arizona}
  \city{Tucson, Arizona}
  \country{United States}
}
\email{tandonr@arizona.edu}

\begin{abstract}
Deep neural networks are vulnerable to adversarial perturbations, limiting deployment in safety-critical applications such as synthetic aperture radar (SAR) automatic target recognition (ATR). Randomized smoothing improves robustness by averaging predictions over noisy inputs, but isotropic noise often fails to preserve the semantic structure of SAR imagery. We propose \emph{semantic smoothing}, a defense that replaces noised-based perturbations with structured randomized transformations generated by a novel view synthesis model. For SAR, we condition on acquisition geometry to synthesize multiple plausible radar views. Predictions across generated randomized views are aggregated to form a robust classifier. Experiments show that semantic smoothing improves robustness against standard attacks, such as FGSM and PGD, and SAR-specific attacks, such as OTSA and SMGAA, while also increasing clean classification accuracy. These results demonstrate that randomized smoothing via semantically preserving geometric transformations is a promising alternative to isotropic noise for adversarial defense in structured sensing domains.
\end{abstract}



\copyrightyear{2026}
\acmYear{2026}
\setcopyright{cc}
\setcctype{by}
\acmConference[GLSVLSI '26]{Great Lakes Symposium on VLSI 2026}{June 22--24, 2026}{Canandaigua, NY, USA}
\acmBooktitle{Great Lakes Symposium on VLSI 2026 (GLSVLSI '26), June 22--24, 2026, Canandaigua, NY, USA}
\acmDOI{10.1145/3787109.3816381}
\acmISBN{979-8-4007-2431-2/2026/06}

\begin{document}

\maketitle

\section{Introduction}
Deep neural networks have achieved remarkable success across a wide range of computer vision tasks, yet their susceptibility to adversarial perturbations remains a major barrier to reliable deployment in safety-critical environments \cite{goodfellow2014explaining, madry2017towards}. Carefully crafted perturbations, often imperceptible to human observers, can cause severe model misclassifications and undermine trust in automated recognition systems. As machine learning continues to be integrated into real-world sensing and decision pipelines, developing effective adversarial defenses is increasingly important.

This challenge is particularly prevalent in synthetic aperture radar (SAR) automatic target recognition (ATR), where classification errors may cause significant operational consequences. Unlike optical imagery, SAR systems actively illuminate a scene with electromagnetic signals and reconstruct images from the returned echoes. As a result, SAR image formation is strongly dependent on acquisition geometry, and object appearance can vary substantially with observation conditions such as azimuth and elevation angles. Recent work has shown that SAR classifiers are vulnerable not only to conventional gradient-based attacks, but also to domain-specific adversarial perturbations that exploit the underlying physics of radar image formation \cite{ye2023realistic, peng2022scattering}. Robust defense strategies are therefore especially important to this setting.

Among existing defenses, randomized smoothing has emerged as a principled and widely studied approach for improving adversarial robustness with provable guarantees \cite{cohen2019certified}. Randomized smoothing constructs a robust classifier by averaging predictions over randomly perturbed copies of an input, typically using isotropic Gaussian noise. This framework offers certified robustness guarantees and has demonstrated effectiveness against many standard noise-based adversaries. However, its perturbation model is agnostic to the structure of the underlying data. For complex sensing modalities such as SAR, additive isotropic noise often produces samples that do not correspond to meaningful observations of the scene. This limitation becomes especially pronounced when facing physically grounded attacks that exploit scattering behavior or acquisition geometry rather than simple pixel-space perturbations.

\begin{figure*}[t]
    \centering
    \includegraphics[width=0.9\textwidth]{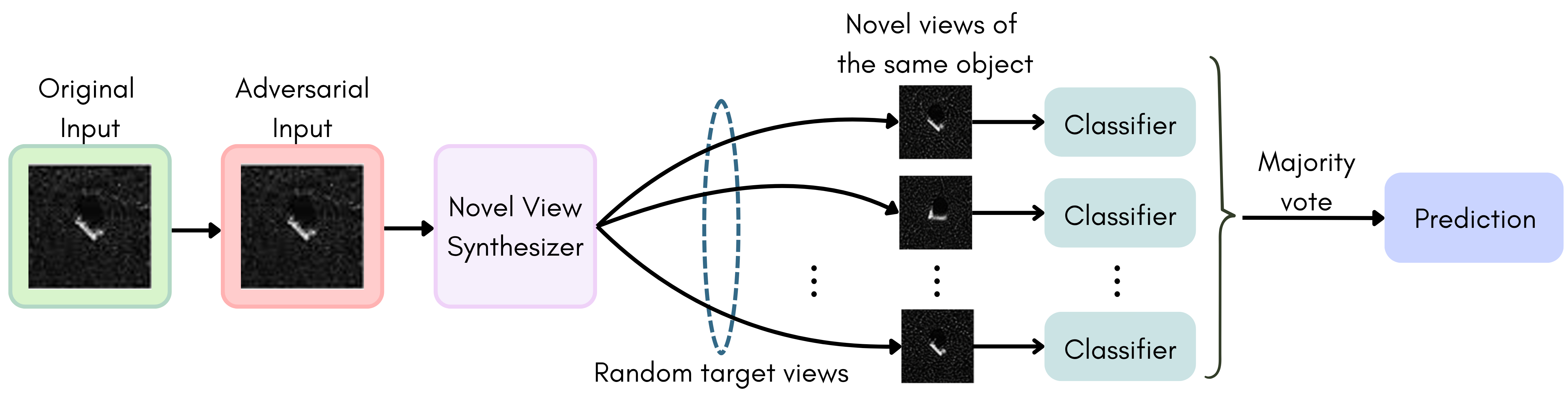}
    \caption{Overview of the proposed semantic smoothing defense. Given a potentially adversarial input, a geometry-conditioned novel view synthesis model generates a set of randomized, semantics-preserving views. Each generated view is evaluated by the classifier, and the final prediction is obtained by aggregating the individual predictions through majority voting.}
    \label{fig:main_fig}
\end{figure*}

\textbf{Motivation for Semantic Smoothing:} These shortcomings motivate the need for smoothing mechanisms that preserve semantics rather than destroy them. Ideally, perturbed samples used for robust aggregation should remain plausible observations of the same underlying object. Recent advances in generative modeling, including diffusion models and neural rendering provide powerful tools for synthesizing realistic images under controlled transformations \cite{chan2022efficient, mildenhall2021nerf, chan2023generative}. However, most existing defenses using generative models focus on reconstruction or purification and do not explicitly exploit structured transformation spaces for robustness \cite{samangouei2018defense, nie2022diffusion}. In SAR, this gap is even more pronounced where few methods leverage the known geometry of the sensing process to produce semantics preserving adversarial defenses.

To address these challenges, we propose \emph{semantic smoothing}, a defense framework that replaces isotropic noise injection with semantically meaningful transformations generated by a novel view synthesis model. Instead of averaging predictions over noisy inputs, our method aggregates predictions across multiple synthesized views of the same object under randomized but controlled geometric transformations. We instantiate this framework using Generative Novel View Synthesis (GeNVS) \cite{chan2023generative}, adapted for SAR imagery through conditioning on azimuth and elevation parameters. This yields a smoothed classifier that produces accurate object identity on the synthesized novel views sampled from the generative model. An overview of this approach is depicted in Figure \ref{fig:main_fig}.



\textbf{Main Contributions:} Our contributions are summarized as follows:
\begin{itemize}
    \item We introduce \emph{semantic smoothing}, a novel adversarial defense that uses randomized generative transformations to produce multiple semantics-preserving views of an input and aggregates classifier predictions across them.
    \item We develop a SAR-specific instantiation of this framework using geometry-conditioned novel view synthesis based on GeNVS, enabling controllable generation of randomized radar views through varying azimuth and elevation parameters.
    \item Extensive experiments on the SAR ATR MSTAR dataset show that semantic smoothing provides substantial robustness gains against both conventional gradient-based attacks and domain-specific SAR attacks. Compared to randomized smoothing, our method achieves stronger adversarial robustness while also improving clean classification accuracy through aggregation over semantically consistent synthesized views.
\end{itemize}

\section{Related Works}

    \textbf{Attacks and Defenses.} Prior work on adversarial robustness had developed along two main directions: increasingly stronger attack methods and defenses designed to counter such attacks \cite{raghunathan2018certified}. 
    Common attacks include the Fast Gradient Sign Method (FGSM) \cite{goodfellow2014explaining} and Projected Gradient Descent (PGD) \cite{madry2017towards}, and in SAR settings more domain-aware methods such as On-Target Scatterer Attack (OTSA) \cite{ye2023realistic} and Scattering Model Guided Adversarial Attack (SMGAA) \cite{peng2022scattering}, which model physically meaningful scatterer perturbations rather than arbitrary pixel noise \cite {ye2023realistic, peng2022scattering}.
    In this line of literature, robustness usually refers to whether a classifier still predicts the correct label after an adversarial perturbation, and results are often reported through quantities such as \textbf{attack success rate} or \textbf{robust accuracy} under a specified threat model \cite{goodfellow2014explaining}. 
    
    On the defense side, the most common approaches include empirical methods such as adversarial training and certifiable defenses, including local Lipschitz based methods \cite{hein2017formal} that bounds the model's output, and randomized smoothing, which aims to provide certified robustness by averaging predictions over noisy copies of the input \cite{cohen2019certified, lecuyer2018certified, rosenfeld2020certified}. 
    There are also prior works using generative models as a defense that purify the adversarial inputs using GANs or diffusion models, such as Defense-GAN \cite{samangouei2018defense} and Diff-Pure \cite{nie2022diffusion}.

\textbf{Novel View Synthesis.} Novel view synthesis (NVS) aims to generate unobserved views of a scene or object from one or more input images. Early approaches relied on explicit geometry and image-based rendering techniques, such as view morphing and multi-view reconstruction, to interpolate physically valid viewpoints from sparse observations \cite{seitz1996view, debevec2023modeling}. More recently, neural scene representation methods have become dominant. Neural Radiance Fields (NeRF) \cite{mildenhall2021nerf} learn continuous volumetric scene representations that can be rendered from arbitrary viewpoints, while extensions such as PixelNeRF \cite{yu2021pixelnerf} condition these representations directly on image inputs to improve efficiency and generalization.

Generative approaches have further advanced NVS by learning scene priors across datasets rather than optimizing each scene independently. GAN-based methods such as GRAF \cite{schwarz2020graf} and EG3D \cite{chan2022efficient} demonstrated high-quality 3D-aware synthesis, while diffusion-based models offer improved training stability and sample fidelity. In particular, GeNVS \cite{chan2023generative} combines conditional diffusion modeling with latent 3D feature rendering to generate geometrically consistent novel views from input images and target poses. We adopt GeNVS in this work due to its ability to model complex data distributions and synthesize realistic, controllable views for robust SAR classification.

\section{Methodology}

Our approach is motivated by the limitations of randomized smoothing as an adversarial defense. While randomized smoothing provides robustness by injecting noise into the attacked input image, the perturbations are typically isotropic and agnostic to the underlying structure of the data. As a result, the smoothing process is confined to a local pixel-level neighborhood of the input and does not explicitly account for whether the resulting samples correspond to plausible variations of the original object.

In contrast, semantic smoothing introduces a semantically grounded form of randomization based on generative modeling. Rather than perturbing attacked inputs with random noise, we use a generative model to produce meaningful variations of the input that preserve the underlying object identity. This changes the nature of the smoothing process: instead of averaging predictions over small, isotropic perturbations around the attacked image, semantic smoothing averages over randomized views that may be substantially different in pixel space while remaining close in semantic content. In this sense, semantic smoothing can move the input away from the adversarial pixel-level neighborhood and toward plausible points on the data manifold. The key intuition is that when an adversarially perturbed image is passed through such a model, the generated views can suppress non-semantic adversarial artifacts while preserving the semantic content needed for classification.

\subsection{Semantic Smoothing via NVS}

\begin{algorithm}[t]
\caption{Semantic Smoothing via Generative Novel View Synthesis}
\label{alg:semantic_smoothing}
\begin{algorithmic}[1]
\Require Input image $x^{adv}$, generative model $G(\cdot)$, classifier $f_{cls}(\cdot)$, transformation distribution $T_D$, number of samples to generate $N$
\Ensure Final predicted label $\hat{y}$

\State Initialize $\mathcal{Y} \gets \emptyset$

\For{$i = 1$ to $N$}
    \State Sample transformation parameters $t_i \sim T_D$
    \State Generate semantic-preserving view $x^{NV}_i = G(x^{adv}, t_i)$
    \State Obtain prediction $y_i = f_{cls}(x^{NV}_i)$
    \State $\mathcal{Y} \gets \mathcal{Y} \cup \{y_i\}$
\EndFor

\State $\hat{y} \gets \arg\max_{c} \sum_{i=1}^{N} \mathbf{1}(y_i = c)$

\State \Return $\hat{y}$
\end{algorithmic}
\end{algorithm}

To apply this idea, we draw inspiration from NVS. NVS models are capable of generating new views of an object by varying parameters such as viewpoint, rotation and perspective while maintaining consistency of the original scene. This provides a natural mechanism for controlled randomization. Instead of adding pixel-level noise, we sample across a space of physically meaningful transformations. By querying the generative model at different viewpoints, we obtain a set of images that reflect the same object under varied but semantically consistent conditions. These generated views serve as the basis for aggregated prediction similar to the Monte Carlo sampling used in randomized smoothing.

We assume access to a conditional generative model that, given an input instance, can produce a distribution of semantically equivalent outputs under controlled randomized transformations. These transformations may correspond to viewpoint changes or geometric perturbations. The key requirement is that the generative process maps inputs back onto the manifold of valid data while maintaining label consistency.
Formally, let $x$ denote the unobserved original input and $x^{adv}$ the observed perturbed input. We define a transformation distribution $T_D$ over a set of semantic-preserving parameters. For each sampled transformation $t_i \sim T_D$, the generative model $G$ produces a novel output $x^{NV}_{i} \sim G(x^{adv}, t_i)$. This results in a collection of synthesized novel views $\{x^{NV}_i \}_{i=1}^N$, each of which is expected to retain the true class semantics while mitigating adversarial artifacts.

Each generated view is then passed through a pretrained classifier $f_{cls}(\cdot)$ producing a set of predictions $\{f_{cls}(x^{NV}_i)\}_{i=1}^N$. The final prediction is obtained via a majority vote over the sampled outputs. Intuitively, while adversarial perturbations may cause misclassifications in the original inputs, they are unlikely to persist consistently across the space of semantic transformations induced by the generative model. This results in features corresponding to the true class remaining stable leading to a consistent consensus among the predictions.

We call this procedure \emph{semantic smoothing} over the transformation space induced by the generative model. We give a procedure for this algorithm in Algorithm \ref{alg:semantic_smoothing}. By replacing isotropic noise with structured, semantics-preserving randomization, we obtain a defense mechanism that is both more interpretable and better aligned with the underlying data distribution.

\begin{figure}
    \centering
    \includegraphics[width=1.0\linewidth]{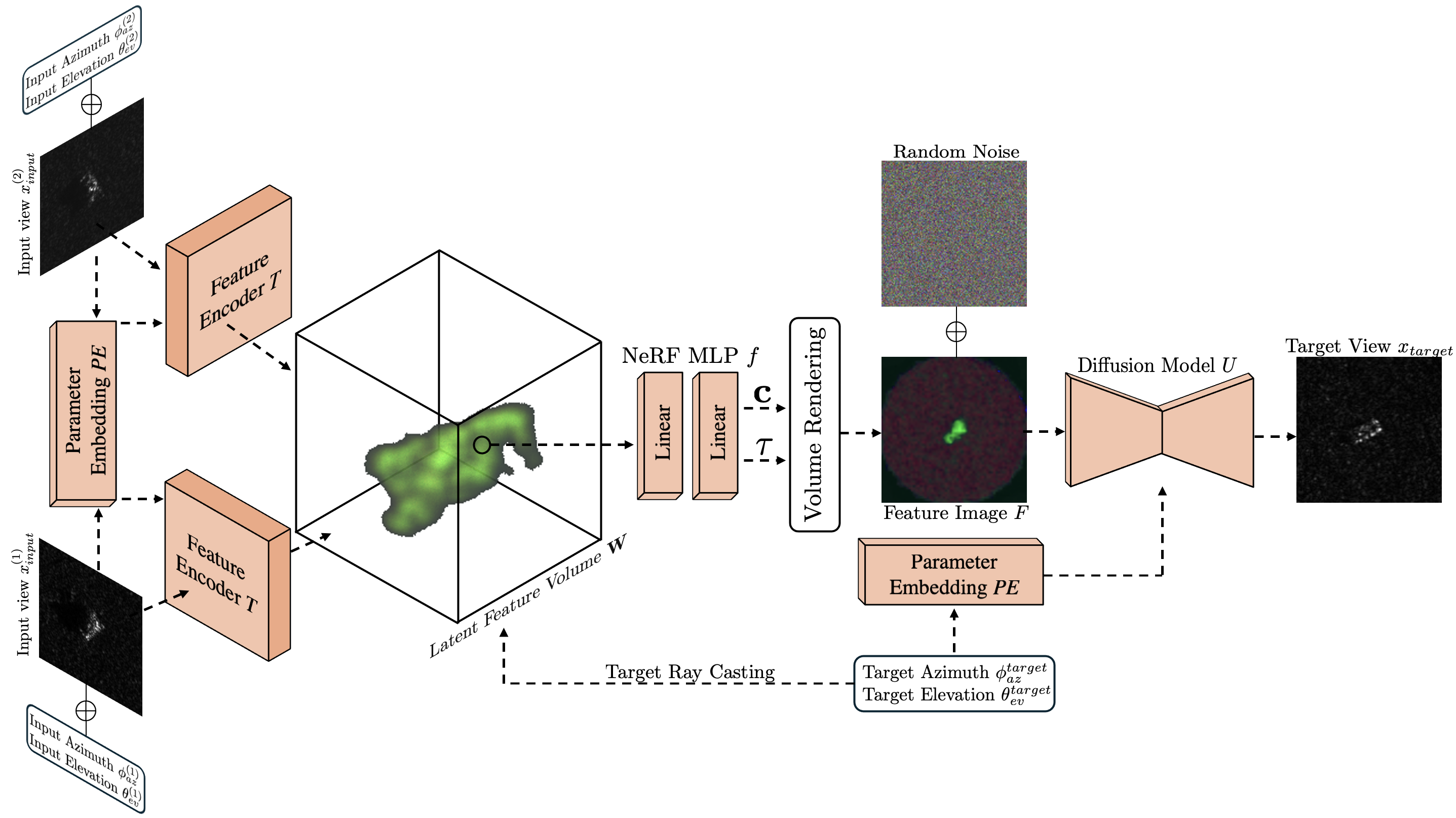}
    \caption{Overview of our SAR image generation pipeline. Given input SAR images and their imaging geometry, the model encodes them into latent features, renders those features from a target viewpoint, and uses a diffusion model to generate the final SAR image at the desired view..}
    \label{fig:gen_fig}
    \vspace{-15pt}
\end{figure}

\subsection{Application to SAR}

\begin{figure*}[t]
    \captionsetup[subfigure]{font=scriptsize}
    \centering
    
    \begin{subfigure}{0.186\textwidth}
        \centering
        \includegraphics[width=\linewidth]{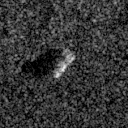}
        \caption{Original Image}
        \label{fig:sub1}
    \end{subfigure}
    \hfill
    \begin{subfigure}{0.186\textwidth}
        \centering
        \includegraphics[width=\linewidth]{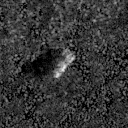}
        \caption{FGSM Attacked Image}
        \label{fig:sub2}
    \end{subfigure}
    \hfill
    \begin{subfigure}{0.186\textwidth}
        \centering
        \includegraphics[width=\linewidth]{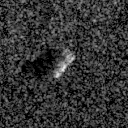}
        \caption{PGD Attacked Image}
        \label{fig:sub3}
    \end{subfigure}
    \hfill
    \begin{subfigure}{0.186\textwidth}
        \centering
        \includegraphics[width=\linewidth]{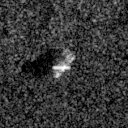}
        \caption{OTSA Attacked Image}
        \label{fig:sub5}
    \end{subfigure}
    \hfill
    \begin{subfigure}{0.186\textwidth}
        \centering
        \includegraphics[width=\linewidth]{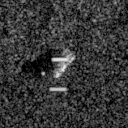}
        \caption{SMGAA Attacked Image}
        \label{fig:sub6}
    \end{subfigure}

    \caption{The original MSTAR image (a) for the 2S1 object category followed by the perturbed image via the generic attacks FGSM (b) and PGD (c) and the SAR-specific attacks OTSA (d) and SMGAA (e).}
    \label{fig:attacks}
\end{figure*}

Adversarial robustness is particularly critical in SAR where automated target recognition (ATR) systems are often deployed in high-stakes setting. Misclassifications in these systems can lead to severe real-world consequences, such as incorrect identification of vehicles in defense applications. Failures in SAR-based ATR can directly impact operational decision-making, making SAR a compelling domain for studying and mitigating adversarial vulnerabilities \cite{11267023, 11234188}.

SAR imaging is fundamentally different from optical imaging. Rather than passively capturing reflected light, SAR systems transmit electromagnetic signals toward a scene and record the returned echoes. By coherently processing these returns over a moving platform, a high-resolution image is formed. A key aspect in this process is that the imaging geometry and the transmitted signal is explicitly known. During acquisition, the system records its relative platform orientation to the ground in terms of azimuth ($\phi_{az}$) and elevation ($\theta_{el}$) angles. These parameters strongly influence the resulting image, giving rise imaging effects such as layover and shadowing \cite{richards2009remote}. Consequently, the appearance of an object in SAR imaging can vary significantly with the viewpoint, even though its semantic identity remains unchanged.

This inherent dependence on viewing geometry makes SAR particularly well-suited for our semantic smoothing framework. Instead of treating variability as isotropic noise, we explicitly model it through physically meaningful transformations with respect to $\phi_{az}$ and $\theta_{el}$.

We define the generative model $G(\cdot)$ as a conditional SAR view synthesis model. Given a clean input image $x$ with parameters ($\phi_{az}^{input}, \theta_{el}^{input}$), we produce an adversarial counterpart $x^{adv}$ by applying some perturbation to $x$. We then generate novel views by sampling target viewing parameters $t_i=(\phi_{az}^{target}, \theta_{el}^{target}) \sim T_D$ where $T_D$ defines the distribution over plausible SAR acquisition geometry. In practice we define $T_D$ to be the uniform distribution over training data SAR acquisition parameters. Each generated view is then given by $x^{NV}_i \sim G(x^{adv}, t_i)$.



\subsubsection{SAR Novel View Generation}

\begin{table*}[t]
\centering
\caption{Robustness accuracy of randomized smoothing and semantic smoothing under standard gradient-based attacks and SAR-specific physically grounded attacks.}
\label{tab:robustness}
\begin{tabular}{ll c ccc ccc}
\toprule
\multirow{2}{*}{\makecell{\textbf{Attack}\\\textbf{Type}}} & \multirow{2}{*}{\makecell{\textbf{Attack}\\\textbf{Strength}}} 
  & \multirow{2}{*}{\textbf{No Def.} }
  & \multicolumn{3}{c}{\textbf{Rand. Smoothing}} 
  & \multicolumn{3}{c}{\textbf{Semantic Smoothing}} \\
\cmidrule(lr){4-6}\cmidrule(lr){7-9}
 & & & \multicolumn{1}{c}{$\sigma = 0.002$} 
       & \multicolumn{1}{c}{$\sigma = 0.01$} 
       & \multicolumn{1}{c}{$\sigma = 0.02$} 
       & \multicolumn{1}{c}{@1} 
       & \multicolumn{1}{c}{@5} 
       & \multicolumn{1}{c}{@10} \\
\midrule
\multirow{1}{*}{No Attack}
  & N/A & 90.00\% & 99.45\% & 28.91\% & 20.00\% & 85.09\% & 92.36\% & 94.72\% \\
\midrule
\multirow{3}{*}{FGSM}
  & $\varepsilon{=}0.001$ & 57.58\% & 64.18\% & 23.64\% & 20.18\% & 84.72\% & 92.90\% & 94.72\% \\
  & $\varepsilon{=}0.002$ & 18.18\% & 18.54\% & 20.36\% & 20.54\% & 84.72\% & 92.18\% & 95.81\% \\
  & $\varepsilon{=}0.005$ & 0.20\%  & 0.54\%  & 12.00\% & 20.09\% & 83.45\% & 91.63\% & 94.36\% \\
\midrule
\multirow{3}{*}{PGD}
  & $\varepsilon{=}0.001$ & 44.04\% & 59.63\% & 24.06\% & 20.36\% & 82.90\% & 90.90\% & 95.45\% \\
  & $\varepsilon{=}0.002$ & 7.07\%  & 11.45\% & 20.36\% & 20.91\% & 84.72\% & 92.54\% & 94.18\% \\
  & $\varepsilon{=}0.005$ & 0.00\%  & 0.00\%  & 10.91\% & 20.55\% & 82.36\% & 91.81\% & 94.18\% \\
\midrule
\multirow{3}{*}{OTSA}
  & 1 scatter  & 59.19\% & 62.91\% & 14.18\% & 17.36\% & 82.45\% & 87.45\% & 90.90\% \\
  & 2 scatters & 33.54\% & 33.82\% & 9.09\%  & 16.55\% & 80.18\% & 86.72\% & 88.00\% \\
  & 3 scatters & 30.51\% & 27.64\% & 7.09\%  & 16.18\% & 75.27\% & 84.54\% & 85.63\% \\
\midrule
\multirow{3}{*}{SMGAA}
  & 1 scatter  & 53.94\% & 60.27\% & 12.18\% & 17.82\% & 81.45\% & 86.00\% & 91.63\% \\
  & 2 scatters & 31.52\% & 28.73\% & 8.55\%  & 17.09\% & 80.72\% & 90.54\% & 91.45\% \\
  & 3 scatters & 27.68\% & 23.27\% & 6.00\%  & 16.36\% & 81.09\% & 86.00\% & 92.72\%  \\
\bottomrule
\end{tabular}%
\end{table*}

We implement $G(\cdot)$ using a modified denoising diffusion probabilistic model (DDPM) \cite{ho2020denoising} combined with a geometry-aware NVS framework. Specifically, we adopt the GeNVS \cite{chan2023generative} approach with tailored conditioning on SAR acquisition parameters to control the desired output viewpoint. This enables sampling from the conditional distribution
\begin{equation}
    p_{\theta}(x^{NV}_i|x^{adv}, \phi^{input}_{az}, \theta^{input}_{el} \phi^{target}_{az, i}, \theta^{target}_{el, i}).
\end{equation}



The generation pipeline is illustrated in Figure \ref{fig:gen_fig}. First, an image-to-image translation network $T(\cdot)$ maps the input image into a latent 3D feature volume $\mathbf{W}$ conditioned on the input acquisition parameters via a learned embedding function $PE(\cdot)$. This embedding encourages the model to capture the physical relationship between different SAR viewpoints. We note that our goal is to generate semantically consistent novel views that mimic the distribution of the real data. Thus, our goal is to train a generative model that can learn the distribution of the data well, not to properly build a generator that closely models the physical process of SAR rendering.

Given a target viewing configuration $t_i=(\phi_{az}^{target}, \theta_{el}^{target})$, we cast rays through the latent volume $\mathbf{W}$. For each sampled point $r$ along a ray, trilinear interpolation is used to extract a feature vector $w=\mathbf{W}(r)$. This feature is then passed through a NeRF-style multilayer perceptron $f(\cdot)$ \cite{mildenhall2021nerf} which outputs a density $\tau$ and feature representation $\mathbf{c}$ (i.e, $(\tau, \mathbf{c} = f(w))$). These outputs are integrated via volume rendering to produce a feature image
\begin{equation}
\begin{aligned}
F(&x^{adv}, \phi_{\text{az}}^{\text{input}}, \theta_{\text{el}}^{\text{input}}, 
   \phi_{\text{az}}^{\text{target}}, \theta_{\text{el}}^{\text{target}}) \\
= \; &\text{RENDER}\big(
    f \circ T(x^{adv}, PE(\phi_{\text{az}}^{\text{input}}, \theta_{\text{el}}^{\text{input}})), \phi_{\text{az,i}}^{\text{target}}, \theta_{\text{el,i}}^{\text{target}}
\big).
\end{aligned}
\end{equation}
This feature image encodes the scene as viewed from the target geometry and serves as a conditioning input to the diffusion model. During inference, Gaussian noise $y$ is iteratively denoised by a network $U(\cdot)$ conditioned on both the rendered feature image and the target parameter embedding as
\begin{equation} \label{eq:denoiser-mod}
    \begin{split}
        x^{NV}_{i} = U(& y, F(x^{adv}, \phi_{az}^{input}, \theta_{el}^{input}, \phi_{az, i}^{target}, \theta_{el, i}^{target}), \\
        & PE(\phi_{az, i}^{target}, \theta_{el,i}^{target})).
    \end{split}
\end{equation}
Through this process, the model learns to synthesize high-fidelity SAR images corresponding to arbitrary target viewpoints while preserving the underlying object semantics. Importantly, adversarial perturbations in $x^{adv}$ are not consistently preserved across generated views as they do not conform to the learned geometric and physical priors of the data distribution.

This formulation naturally extends to multiple input views by constructing separate latent volumes $\mathbf{W}_j$ for each input $x^{adv}_j$ and averaging the sampled features along each ray before applying the NeRF and rendering steps. The remainder of this pipeline remains unchanged.

\textbf{Adapting Algorithm \ref{alg:semantic_smoothing} for SAR:} Given the adversarial input $x^{adv}$ with original acquisition parameters $(\phi_{az}^{input},\theta_{el}^{input})$, we first sample the transformation parameters $t_i = (\phi_{az, i}^{target}, \theta_{el, i}^{target}) \sim T_D$. We then sample $N$ novel views $\{x_i^{NV}\}_{i=1}^N$ from the generative model $p_{\theta}(x^{NV}_i|x^{adv}, \phi^{input}_{az}, \theta^{input}_{el} \phi^{target}_{az, i}, \theta^{target}_{el, i})$. Each $x_i^{NV}$ is then passed to the pretrained classifier $f_{cls}(x_i^{NV})$ to obtain a prediction $y_i$. After all $N$ views have been evaluated, the final class label is determined through majority voting over the set of predictions $\{y_i\}_{i=1}^N$.




\section{Results and Discussion}

\textbf{Dataset.} We test our semantic smoothing procedure on a subset the publicly released version of the MSTAR dataset \cite{ross1998standard}, one of the most commonly used datasets in SAR ATR. This subset of the MSTAR dataset contains 50 randomly selected SAR image chips for each of ten ground vehicle classes under a variety of imaging conditions for a total of $500$ images. Each chip is distributed in Phoenix binary format, which includes a fixed-length ASCII header followed by the image data. The header encodes collection parameters such as platform geometry, measured azimuth and elevation angles, and target type, while the binary block contains the SAR image matrix. The standard chip size is $128 \times 128$ pixels stored as complex values from which standard 8-bit images can be formed.




\textbf{Attack Methods.} To evaluate the effectiveness of semantic smoothing, we consider a diverse set of adversarial attack methods spanning both generic and SAR-specific threat models. Given the clean input $x$ with ground-truth label $y_{gt}$ and a pretrained classifier $f_{cls}(\cdot)$, each attack constructs an adversarial example $x^{adv}$ in the following manners. We demonstrate the resulting image of each attack in Figure \ref{fig:attacks}.

\textbf{Generic Attacks:} FGSM \cite{goodfellow2014explaining} is a single-step, gradient-based attack that perturbs the input in the direction that maximally increases the classification loss. The magnitude of the perturbation is controlled by a parameter $\varepsilon$, making FGSM a fast but relatively weak adversary. PGD \cite{madry2017towards} is an iterative extension of FGSM that applies multiple small gradient-based updates while constraining the perturbation within a fixed $\varepsilon$-bounded region around the original input. Due to its iterative nature, PGD is considered a strong first-order adversary and a standard benchmark for evaluating robustness.

\textbf{SAR Specific Attacks:} OTSA \cite{ye2023realistic} is a SAR-specific attack that generates adversarial examples by introducing physically plausible scattering perturbations. Rather than arbitrary pixel noise, OTSA optimizes the placement and characteristics of additional scatterers such that the resulting image induces misclassifications while remaining consistent with realistic SAR scattering behavior. SMGAA \cite{peng2022scattering} is a physics-guided attack that perturbs underlying scene scattering parameters instead of directly modifying pixel values. These perturbed parameters are passed through a differentiable SAR forward model to produce adversarial images, ensuring that the resulting perturbations correspond to physically realizable changes in the scene.

\textbf{Randomized Smoothing.} We evaluate our semantic smoothing defense to the commonly used defense of randomized smoothing \cite{cohen2019certified}. Randomized smoothing constructs a robust classifier by averaging predictions over noisy perturbations of the input. Given a base classifier $f_{cls}(\cdot)$, the smoothed classifier $h(\cdot)$ is defined as
\begin{equation}
    h(x) = \underset{c}{\operatorname{argmax}} \quad \mathbb{P}_{z \sim \mathcal{N}(0, \sigma^2I)} (f_{cls}(x + z)=c)
\end{equation}
where $z$ is isotropic Gaussian noise with variance $\sigma^2$. In practice, this expectation is approximated via Monte Carlo sampling by evaluating the classifier on multiple noisy realizations of the input and selecting the majority prediction. The parameter $\sigma$ controls the trade-off between robustness and accuracy, with larger values providing stronger smoothing at the cost of fidelity. Randomized smoothing has the desirable property of enabling certified robustness guarantees, but its reliance on isotropic noise often leads to samples that do not preserve the semantic structure of the input, motivating the semantically grounded alternative proposed in this work.

\vspace{-10pt}
\subsection{Results}

Table \ref{tab:robustness} summarizes the robustness performance of the pretrained classifier, randomized smoothing, and the proposed semantic smoothing approach under a range of adversarial attacks and strengths. When using semantic smoothing, $@N$ refers to the number of novel images generated that are then fed to a classifier proceeded by a majority vote. We report the results of randomized smoothing in Table \ref{tab:robustness} using 100 noised samples that are then aggregated for prediction. From this, several clear trends emerge.

\textbf{Consistent Robustness Gains from Semantic Smoothing:} Semantic smoothing consistently and substantially improves robustness across all evaluated attack methods. Under FGSM and PGD, where the undefended model rapidly degrades as the perturbation strength $\varepsilon$ increases, semantic smoothing maintains high accuracy across all settings. Improvements extend beyond standard attacks to the more structured, domain-specific SAR perturbations. Both OTSA and SMGAA enforce physically consistent scattering behavior in the adversarial inputs, yet semantic smoothing is still able to mitigate adversarial scatters. This demonstrates that the method is robust not only to pixel-level perturbations but also to physically grounded adversarial manipulations.

\begin{table}[t]
\centering
\footnotesize
\caption{Robustness comparison of randomized smoothing and semantic smoothing under equal sampling budget $@N$.}
\label{tab:complexity}
\begin{tabular}{llcccccc}
\toprule
\multirow{2}{*}{\makecell{\textbf{Attack}\\\textbf{Type}}} &
\multirow{2}{*}{\makecell{\textbf{Attack}\\\textbf{Strength}}} &
\multicolumn{3}{c}{\makecell{\textbf{Randomized}\\\textbf{Smoothing} ($\sigma=0.002$)}} &
\multicolumn{3}{c}{\makecell{\textbf{Semantic}\\\textbf{Smoothing}}} \\

\cmidrule(lr){3-5}\cmidrule(lr){6-8}

& &
\textbf{@1} & \textbf{@5} & \textbf{@10} &
\textbf{@1} & \textbf{@5} & \textbf{@10} \\

\midrule

\multirow{2}{*}{FGSM}
& $\varepsilon = 0.001$ & 63.45\% & 63.45\% & 62.91\% & 84.72\% & 92.90\% & 94.72\% \\
& $\varepsilon = 0.005$ & 0.54\%  & 0.54\%  & 0.55\% & 83.45\% & 91.63\% & 94.36\% \\

\midrule

\multirow{2}{*}{PGD}
& $\varepsilon = 0.001$ & 45.09\% & 45.09\% & 45.09\% & 82.90\% & 90.90\% & 95.45\% \\
& $\varepsilon = 0.005$ & 0.00\%  & 0.00\%  & 0.00\% & 82.36\% & 91.81\% & 94.18\% \\

\bottomrule
\end{tabular}
\vspace{-20pt}
\end{table}

\textbf{Semantic Smoothing can Improve Clean Accuracy:} We observe that semantic smoothing also improves the performance in the absence of adversarial perturbations. On clean inputs, we see accuracy increase when aggregating multiple generated views. This indicates that the method does not merely act as a defense, but also enhances generalization by leveraging semantically consistent transformations of the input.

\textbf{Semantic Smoothing Outperforms Randomized Smoothing:} When compared to randomized smoothing, we observe a more nuanced trend. Randomized smoothing performance varies depending on the choice of perturbation radius $z$. For certain $z$ values, modest improvements are observed, but performance is inconsistent and often degrades for other choices. This sensitivity highlights a key limitation of isotropic noise-based smoothing, as the lack of semantic structure in the perturbations can lead to samples that do not preserve class-relevant information. In contrast, semantic smoothing provides stable and significant improvements across all settings without requiring careful tuning of a noise scale parameter.

Table \ref{tab:complexity} compares randomized smoothing and semantic smoothing under an equal sampling budget where both methods use the same number of classifier evaluations $N$. Under this setting, randomized smoothing is computationally simpler requiring only the classification of $N$ noisy samples yielding an overall cost proportional to $N\times C_f$, where $C_f$ denotes the classifier inference cost. Semantic smoothing incurs additional overhead because each of the $N$ samples must first be synthesized before classification, resulting in a cost proportional to $N \times (C_g + C_f)$, where $C_g$ is the cost of generation. Despite this higher cost, semantic smoothing consistently provides substantially stronger robustness than randomized smoothing when both methods are evaluated with the same value of $N$. These results suggest that the additional complexity of semantic smoothing yields a favorable robustness–computation trade-off, as the generated samples are semantically meaningful and therefore significantly more informative than isotropic noise perturbations.

Overall, these results demonstrate that replacing isotropic noise with structured generative transformations yields a more effective and reliable defense mechanism. Semantic smoothing not only restores robustness under strong adversarial attacks, but also improves performance on clean data, highlighting the benefits of leveraging semantically meaningful variability.

\section{Conclusion}
We introduce \emph{semantic smoothing}, a generative framework that replaces the isotropic noise of randomized smoothing with randomized structured, semantics-preserving transformations. For SAR ATR, we instantiate this framework with a geometry-conditioned NVS model that generates multiple plausible target views and aggregates classifier predictions. Experiments show that semantic smoothing substantially improves robustness against both standard gradient-based and SAR-specific attacks while also improving clean classification accuracy. These results highlight the benefits of smoothing over meaningful transformations rather than arbitrary noise. Future work should study perturbation propagation through GeNVS, evaluate generator robustness, and consider adaptive white-box attacks where the adversary has full knowledge of the defense and optimizes perturbations accordingly.
\vspace{-9pt}


\end{document}